\documentclass[10pt,twocolumn,letterpaper]{article}

\usepackage[pagenumbers]{wacv} 

\usepackage{graphicx}
\usepackage{amsmath}
\usepackage{amssymb}
\usepackage{booktabs}
\usepackage{color}
\usepackage{colortbl}
\usepackage{multirow}
\usepackage{bm}
\usepackage[rightcaption]{sidecap}

\definecolor{wacvblue}{rgb}{0.21,0.49,0.74}

\definecolor{silver}{gray}{0.95}
\definecolor{lightgray}{gray}{0.98}
\definecolor{lightred}{rgb}{0.99,0.94,0.95}
\definecolor{lightblue}{rgb}{0.94,0.95,0.99}
\definecolor{lightgreen}{rgb}{0.94,0.99,0.95}
\definecolor{best}{rgb}{0.62,0.58,0.99}
\definecolor{secondbest}{rgb}{0.84,0.83,0.99}

\usepackage[pagebackref,breaklinks,colorlinks,allcolors=wacvblue]{hyperref}

\def\wacvPaperID{****} 
\def\confName{CV4EO-WACV}
\def\confYear{2027}

\title{Revisiting the Efficient Architectures for Remote Sensing Change Detection}

\author{Mubashir Noman$^{1}$, Mustansar Fiaz$^{2}$, Abdul Hannan$^3$, Hiyam Debary$^{2}$, Shah Nawaz$^4$\\ Fahad Shahbaz Khan$^1$, Salman Khan$^1$\\
$^1$MBZUAI \hspace{2mm} $^2$IBM Research \hspace{2mm} 
$^3$Fondazione Bruno Kessler, Italy \hspace{2mm} \\
$^4$Johannes Kepler University Linz, Austria \hspace{2mm} 
}

\begin{document}
\maketitle

\begin{abstract}
Being a cornerstone of temporal analysis, change detection has been playing a pivotal role in modern earth observation. Existing change detection methods rely on the Siamese encoder to individually extract temporal features followed by temporal fusion. 
Subsequently, these methods design sophisticated decoders to improve the change detection performance without taking into consideration the complexity of the model. 
These aforementioned issues intensify the overall computational cost as well as the network's complexity which is undesirable. 
Alternatively, few methods utilize the early fusion scheme to combine the temporal images. These methods prevent the extra overhead of Siamese encoder, however, they also rely on sophisticated decoders for better performance. In addition, these methods demonstrate inferior performance as compared to late fusion based methods. 
To bridge these gaps, we introduce a simple and effective method for the change detection task. The proposed method performs the early fusion of the temporal data and replaces the decoder with a parameter-free multiscale feature fusion module thereby significantly reducing the overall complexity of the model. Our method demonstrates the optimal balance between change detection performance and the prediction speed across a variety of encoder architectures.
Extensive experimentation on four challenging change detection datasets reveals the effectiveness of the proposed method. Code is available at: \url{https://github.com/techmn/efenet}.
\end{abstract}


\section{Introduction}
\label{sec:intro}
Remote sensing acquires a large amount of temporal data, allowing researchers to study and monitor the earth surface over the passage of time. Being a critical task to investigate changes between temporal data, change detection (CD) has a wide range of applications, including urban planning, deforestation monitoring, disaster analysis, emergency planning, and more. Earlier approaches performed change detection by utilizing change vector analysis \cite{bovolo2011framework} and support vector machines \cite{volpi2013supervised}. 
Later, with the rise of deep learning, convolutional neural network (CNN) based approaches have reshaped the problem of CD task. Few methods \cite{daudt2018fully, jiang2022wricnet, peng2019end} perform the early fusion of the temporal images and utilize an end-to-end network for change detection. 
These networks are typically an encoder-decoder segmentation network with the exception that the input is concatenated bitemporal images.
The early fusion networks allow for direct interaction of temporal data during feature extraction and provide somehow fair change detection performance, however, these approaches are underexplored by the remote sensing community. 

\begin{figure}[t]
\centering
 \includegraphics[width=0.95\linewidth]{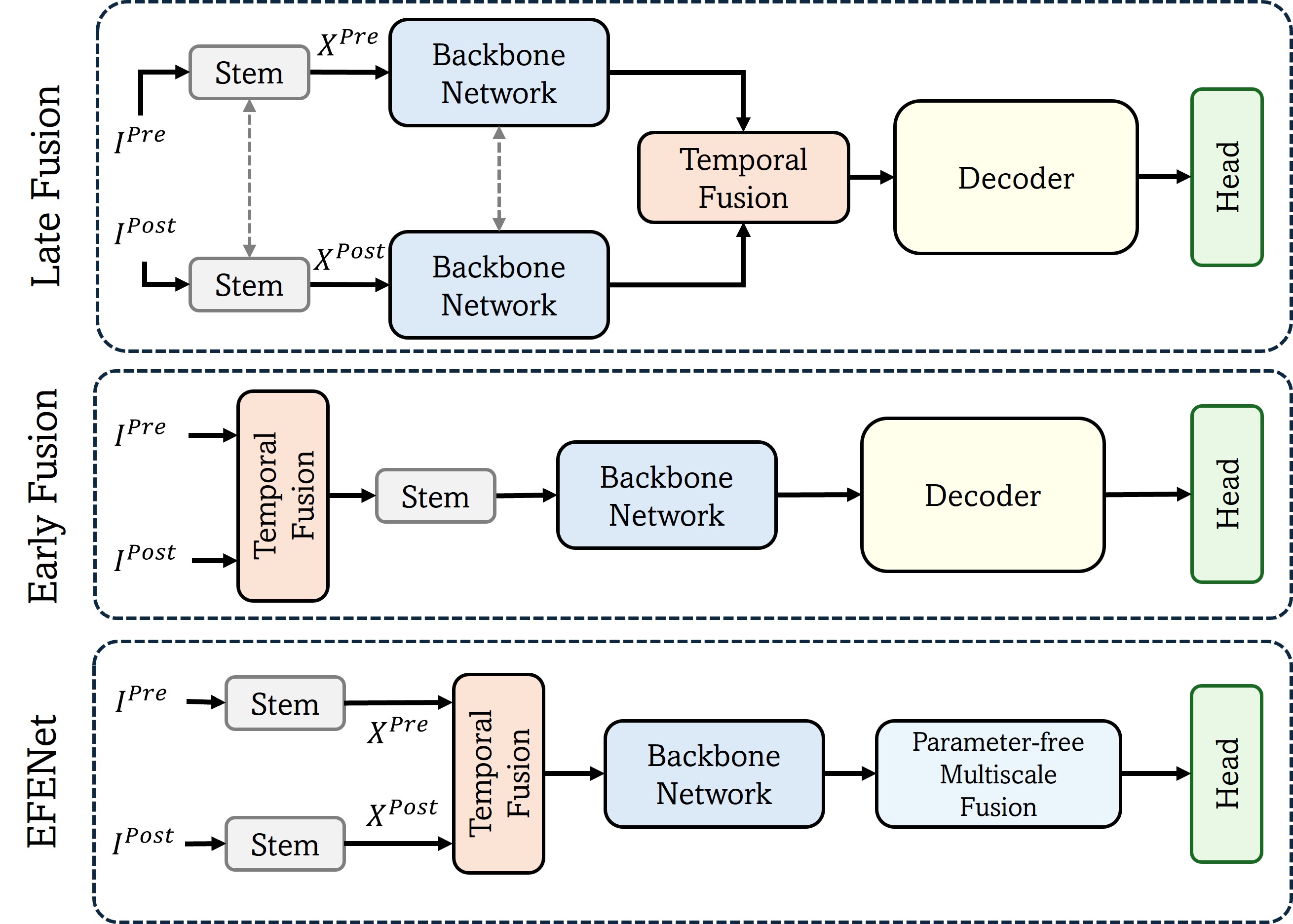} 
 
\caption{Comparison of various CD frameworks. a) In late fusion, $I_{pre}$ and $I_{post}$ are fed to Siamese encoder causing the backbone network to process each image separately which lead to increased computational cost. b) Early fusion approach prevent this by concatenating the bitemporal images before passing it to backbone, however, the sophisticated decoder still adds undesirable complexity to the model. c) Proposed EFENet introduces a simple design that bypasses the extra overhead of Siamese encoder and sophisticated decoder.
}
\vspace{-3mm}
\label{fig:early_vs_late_fusion}
\end{figure}

Alternatively, most of the CNN- and transformer-based CD methods \cite{zhang2020deeply, bandara2022changeformer, 2024_elgcnet, xie2025fsg} fall into the late fusion category. 
As illustrated in Figure~\ref{fig:early_vs_late_fusion}, a typical CD approach that uses the late fusion scheme is composed of a Siamese encoder (shared stem and backbone), a temporal fusion module, a decoder, and a CD head. The Siamese encoder is responsible for extracting the features from bitemporal images one at a time, and the sophisticated decoder combines the multiscale features and highlights the relevant information necessary for CD task. Both these components usually add computational burden to the overall model's performance. 
Although late fusion approaches \cite{li2022transunetcd, noman2024remote, zhao2024rs} demonstrate promising CD performance, it is achieved at the expense of a significant increase in computational complexity.

To address the above limitations of late fusion CD models, we propose a simple yet efficient CD method that takes the advantages of early fusion and replaces the sophisticated decoder with parameter free multi-scale feature fusion module thereby achieving the optimal CD performance. Unlike traditional early fusion approaches, proposed method projects the bitemporal images to the feature space by using stem layer followed by fusion of bitemporal features as illustrated in Figure~\ref{fig:early_vs_late_fusion}. Notably, proposed method does not utilize a sophisticated decoder enabling it to further reduce the computational cost while maintaining the balance between the performance and efficiency. 
We summarize our contributions as follows:

\begin{itemize}
    \item We propose an effective change detection model (EFENet) that is based on early fusion strategy and does not utilize the sophisticated CD decoder while achieving promising CD performance.
    \item We introduce an effective multiscale feature fusion (EMFF) module that does not contain any learnable parameters and emphasizes the semantic information in the multiscale representations, thereby providing optimal CD performance.
    \item We demonstrate how the performance of change detection is affected by using various encoder networks.
    \item Extensive experimentation on four challenging CD datasets with various encoder networks demonstrate the optimal performance revealing the effectiveness of the proposed method.
\end{itemize}

\section{Related Work}
\subsection{Early and Late Fusion Methods}
The advancement of deep learning in remote sensing change detection (RSCD) tasks can be roughly categorized as early fusion and late fusion. The early fusion approaches first combine the bitemporal input images and process them through a single encoder-decoder segmentation network \cite{alcantarilla2018street, peng2019end, de2020change}. Alcantarilla et al. \cite{alcantarilla2018street} first concatenate the two images into six-channel input and pass it to FCN to realize the street view CD. Peng et al. \cite{peng2019end} used UNet++ model with dense skip connections to learn multiscale and various levels of feature representations.  Jiang et al. \cite{jiang2022wricnet} propose a weighted scale block, which dynamically assigns appropriate weights to multiscale features to obtain accurate detection.

Late fusion approaches utilize a shared backbone to extract high-level features for the bitemporal input features and fuse them by employing different operations such as subtraction, multiplication, or concatenation for the change detection task \cite{daudt2018fully, chen2020dasnet}. 
Among the convolutional neural network (CNN) based CD approaches, Zhang et al. \cite{zhang2020feature} used transfer learning to form a two-channel network with shared weights to extract multiscale and multi-depth features using a change magnitude guided loss function. Fang et al. \cite{fang2021snunet} proposed SNUNet-CD, a densely connected Siamese network for comprehensive interaction between the input bitemporal images for change detection. DTCDSCN \cite{liu2020building} and IFN \cite{zhang2020deeply} exploit channel and spatial attention mechanisms to learn the spatially robust features. 
These approaches aim to extract both low-level features to assist in target localization as well as high-level features to enrich the semantic information for better change detection.
Zhang et al. \cite{zhang2021escnet} introduced the superpixel enhanced CD network (ESCNet), which takes advantage of superpixel segmentation to reduce latent noise in pixel-level feature maps while preserving edges, and two weight-sharing superpixel sampling networks (SSNs) tailored for feature extraction.
Lv et al. \cite{lv2022spatial} proposed to integrate an adaptively generated change magnitude image (CMI), which guides the CD model to capture the shape and sizes of the changing regions. 
Huang et al. \cite{huang2024spatiotemporal} proposed a spatiotemporal difference enhancement and adaptive context fusion to aggregate feature maps from multiple scales. 
Huang et al. \cite{huang2021multiple} introduce the MASNet CD method based on selective fusion of multi-temporal features by constructing selective convolutional kernels and multiple attention modules.

\begin{figure*}[t]
\centering
\includegraphics[width=1.0\linewidth]{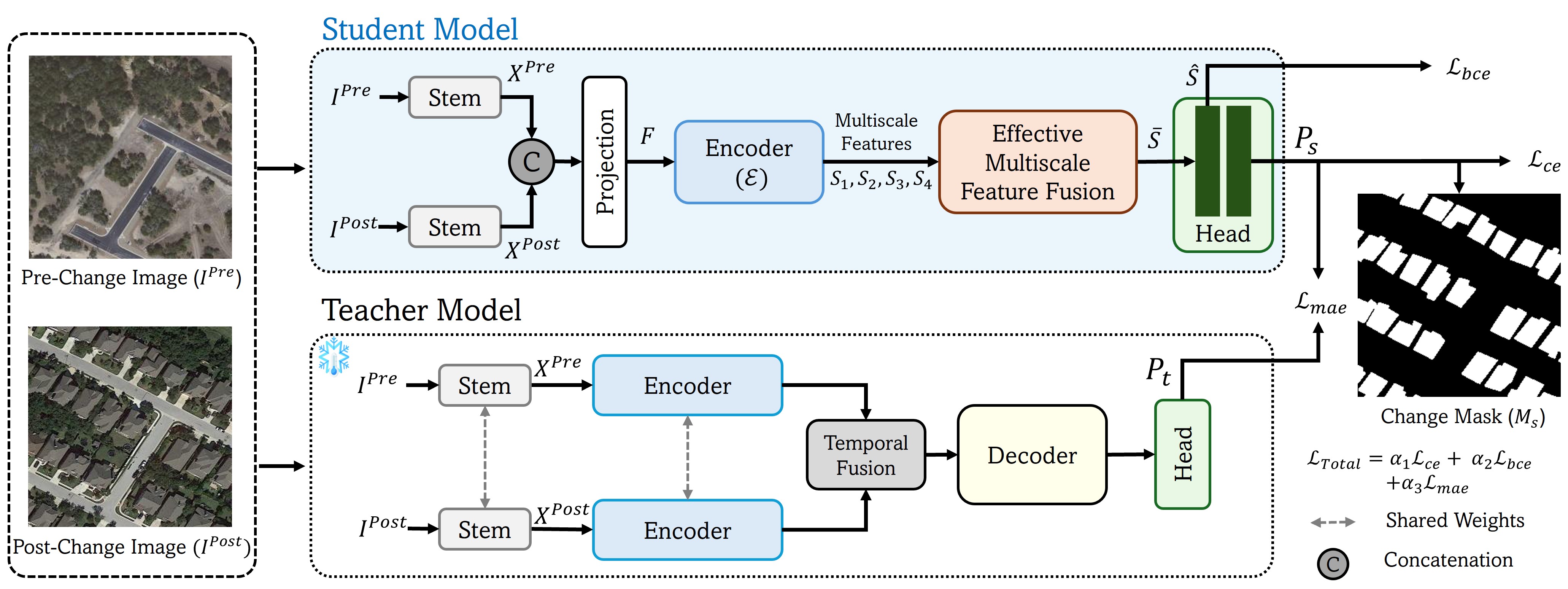} 

\caption{Overall architecture of the proposed EFENet that characterizes a student-teacher framework. The teacher is a large CD network that is used for knowledge distillation. The student network performs early fusion of temporal images and optimally combines the multiscale representations, thereby significantly improving the efficiency of the network.
}
\vspace{-2mm}
\label{fig:main_framework}
\end{figure*}

Additionally, transformer based late fusion CD methods are extensively explored by remote sensing community. 
Bandara et al. \cite{bandara2022changeformer} introduced ChangeFormer, which leverages the vision Transformer as a backbone for extracting bitemporal image features. ScratchFormer \cite{noman2024remote} employed a shuffled sparse-attention operation, and aims to select sparse informative regions to capture the inherent characteristics of the CD data. 
Zhang et al. \cite{zhang2022swinsunet} employed SwinTransformer \cite{liu2021swin} to extract multi-scale features. Liu et al. \cite{liu2022cnn} utilized deep supervision to exploit the multi-scale features and performed multi-scale supervision. 
ELGCNet \cite{2024_elgcnet} employed a hybrid network that encodes global context and local spatial information through a pooled-transpose (PT) attention and depthwise convolution. Li et al. \cite{li2022transunetcd} proposed TransUNetCD, which integrates UNet and Transformer within an encoder-decoder framework. 
Zhang et al. \cite{zhang2024bifa} introduced a CD method called BIFA, which incorporates spatial-temporal alignment and multiscale alignment. Zhang et al. \cite{zhang2025strobustnet} proposed spatial–temporal robust representation learning by enhancing model efficiency and reducing inconsistent bitemporal features confusion. 
Furthermore, mamba \cite{gu2024mamba} based architectures were recently explored in CD domain. 
Zhao et al. \cite{zhao2024rs} proposed RSMamba, which utilizes a selective scan module to encode the global context information in multiple directions, capturing large spatial features from various directions. 
Chen et al. \cite{chen2024changemamba} introduced ChangeMamba, a Mamba-based architecture, to fully learn spatio-temporal features.
CDMamba \cite{zhang2025cdmamba} leverages from state space models and CNNs to dynamically combine global and local features for the CD task.

These aforementioned approaches utilize Siamese encoder network, temporal fusion, a decoder, and a CD head. Although a shared Siamese network extracts meaningful features, it increases the computational complexity. Furthermore, a dedicated decoder is effective for improving CD performance, however, it may result in an undesired increase in computational requirements and model parameters. 
In contrast, we intend to propose a simple but elegant framework that mitigates the requirement of a Siamese encoder and a sophisticated decoder, while  discriminating the complex change regions with better accuracy.

\subsection{Lightweight CD Methods}
Several attempts have been made to improve the efficiency of Siamese-based CD methods. 
Daudt et al. \cite{daudt2018fully} proposed a lightweight method named FC-Siam-Diff, which utilizes a Siamese CNN to fuse bitemporal images along the channel dimension and process them for the CD task.  
Chen et al. \cite{bit2022} proposed a bitemporal image transformer (BIT), which utilizes ResNet18 \cite{he2016resnet} to extract features, followed by a transformer encoder to capture global contextual relationships for change detection. 
Lei et al. \cite{lei2023ultralightweight} proposed lightweight CD networks that utilize an attention operation to recalibrate feature channels for feature enhancement. 
Li et al. \cite{li2023lightweight} used MobileNetV2 backbone pretrained on ImageNet for feature extraction and designed neighbor aggregation module and progressive change identifying module for improved CD performance. 
Han et al. \cite{han2022lwcdnet} introduced the artificial padding convolution (APC) module to improve the coding and decoding ability of features, and designed a new loss function for the bitemporal CD task. 
TinyCD \cite{codegoni2023tinycd} utilized pretrained EfficientNet backbone for feature extraction and attention based feature mixing for efficient CD. 
CDKD \cite{cdkd_2024} utilized the knowledge distillation strategy to improve the performance of lightweight CD methods.
Meng et al. \cite{EDGNet_2026} proposed a lightweight frequency-aware hybrid architecture for practical CD task. 
The lightweight Siamese-based CD methods mainly rely on an efficient encoder to reduce computational load, while the sophisticated decoders are used for better CD performance. Alternatively, we propose an effective early fusion based network that does not use the sophisticated decoder and is able to provide the strong balance between the CD performance and efficiency.

\section{Method}
\label{sec:method}

\subsection{Preliminaries}
\label{ssec:baseline}
We denote the bitemporal images acquired at different time intervals as $I^{pre} \in \mathbb{R}^{3 \times H \times W}$ and $I^{post} \in \mathbb{R}^{3 \times H \times W}$, where $I^{pre}$ is acquired earlier in time than $I^{post}$. Given a pair of bitemporal images $p=(I^{pre}, I^{post})$, the objective of change detection is to highlight the semantic changes between the image pair and predict a binary mask $\mathcal{M} \in \mathbb{R}^{H \times W}$. 

\noindent\textbf{Baseline Framework: }
Considering the efficiency of the model is vital, we adopt the early fusion approach as a baseline framework. The baseline framework takes a bitemporal image pair $(I^{pre}, I^{post})$ and passes it to separate stem layers extracting the feature representations $X^{pre}$ and $X^{post}$. These feature representations are concatenated, fed to convolution layers, and passed to an encoder $\mathcal{E}$. The obtained multiscale feature maps from the encoder are interpolated, then concatenated along the channel dimension, and passed to a convolution layer followed by input to the segmentation head to predict the change map. 

The early fusion in the baseline framework significantly reduces the computations since it eliminates a full encoder pass; however, it also considerably degrades the performance of the model which is undesirable. Additionally, naive replacement of the decoder with a convolution layers further limits the ability of the model to perform favorably for change detection task. 
To this end, we designed an effective change detection network that replaces the decoder with an effective multi-scale feature fusion module having no learnable parameters while exploiting the benefits of early fusion for promising change detection performance. We describe this approach in detail as follows.

\begin{figure}[t]
\centering
\includegraphics[width=0.9\linewidth]{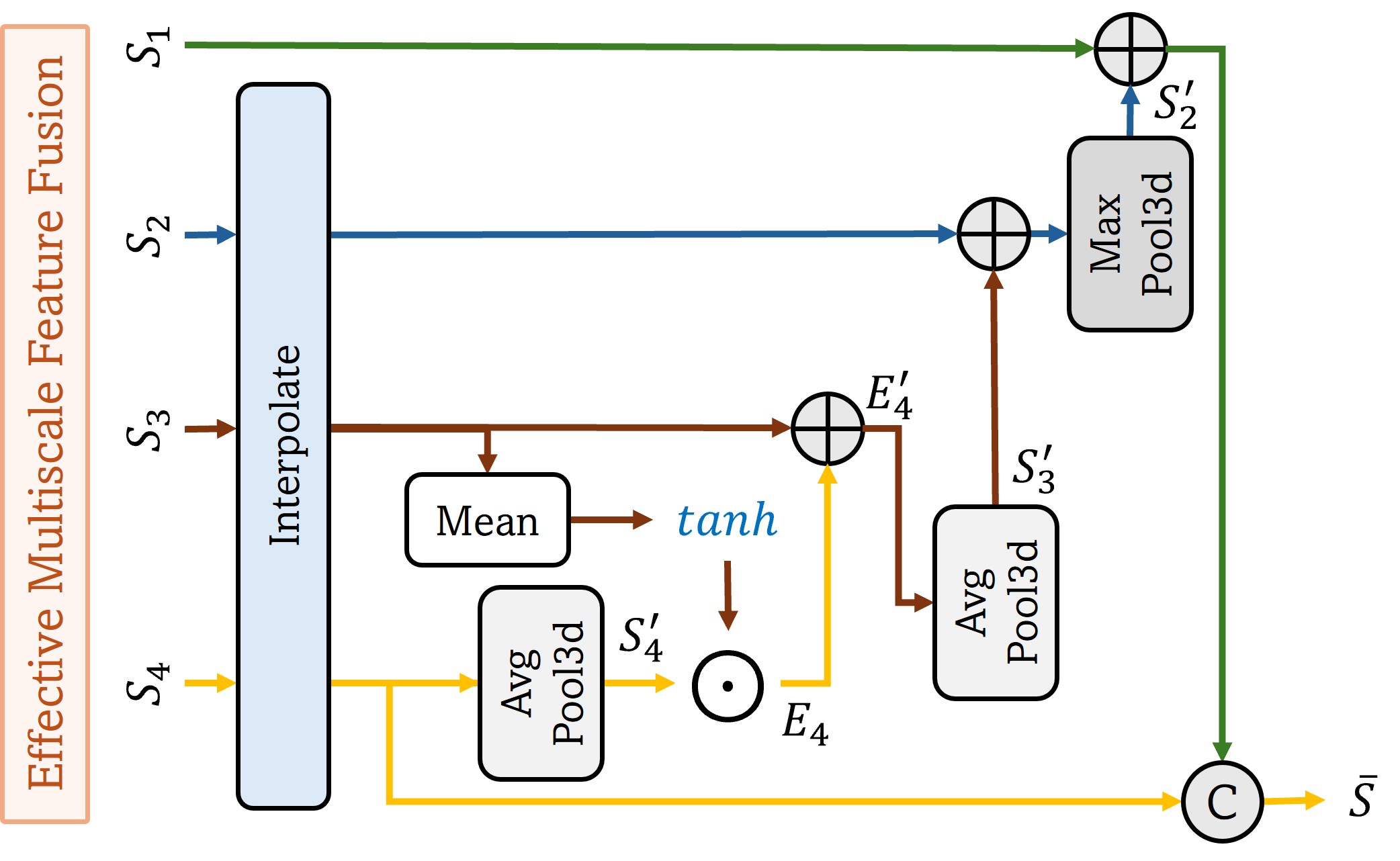}
\caption{Architecture of effective multiscale feature fusion (EMFF) module. EMFF module does not utilize any learnable parameters layer and use pooling operations along the channel dimension to effectively combine the multiscale features while reducing the computational cost. }
\label{fig:emff_arch}
\end{figure}

\subsection{EFENet Framework}
As illustrated in Figure~\ref{fig:main_framework}, the proposed architecture is a distillation-based teacher-student framework. The teacher network is a large CD model composed of a Siamese encoder, a bitemporal and multi-scale feature fusion module, a sophisticated decoder, and a CD head. 
Despite the fact that the student network utilizes early fusion of temporal features, its effective design enables it to perform favorably on the change detection task. The student network consists of an encoder, an effective multi-scale feature fusion (EMFF) module having no learnable parameters, and a CD head. 
The objective of the frozen teacher network is to enable the student network to unlock its full potential.
In other words, the distillation framework instills semantic information from the teacher CD network to the simple student network for explicit enhancement of student representations. 

\subsubsection{Student Network Architecture } Given a bitemporal image pair $p=(I^{pre},I^{post})$, we pass $I^{pre}$ to the stem layer to obtain features $X^{pre}$. 
Similarly, we feed $I^{post}$ to the stem layer to obtain features $X^{post}$. Afterwards, we concatenate the features $X^{pre}$ and $X^{post}$ followed by input to the convolution layers obtaining the fused temporal representations $F$. Mathematically, these operations are represented as:

\begin{equation}
    \begin{array}{l}
    X^{pre}=Conv_{pre}(I^{pre}),  \quad
    X^{post}=Conv_{post}(I^{post}) \\
    X=Concat(X^{pre},X^{post})\\
    F=DWConv(Conv(DWConv(X)))
    \end{array}
\end{equation}

Consequently, we input the feature representations $F$ into the encoder $\mathcal{E}$ to obtain the multi-scale features $S_j$ where $j \in [1,2,3,4]$ represents the $j$th stage of the encoder. 
These multiscale features $S_j$ are effectively fused by the EMFF module to obtain semantically enhanced features $\bar{S}$. Finally, the fused representations are passed to the change detection head that provides the change probabilities $P_s$ to obtain the binary change mask $\mathcal{M}_s$.

\begin{table*}[t!]
\centering
\caption{Summary of remote sensing change detection (RSCD) datasets and their attributes i.e., number of image pairs, image sizes, spatial resolution, splits, image types, and change types.}
\label{tab:cd_datasets}
\scalebox{0.77}{
\begin{tabular}{lcccccc}
\hline
\textbf{Dataset} & \textbf{Image Pairs} & \textbf{Image Size} & \textbf{Spatial Resolution} & \textbf{Train/Val/Test} & \textbf{Image Type} & \textbf{Change Type} \\
\hline
LEVIR-CD \citep{chen2020spatial} & 637 & 1024 $\times$ 1024 & 0.5 m/pixel & 7120 / 1024 / 2048 & satellite images & building \\
CDD-CD \citep{lebedev2018change} & 11 & 4725 $\times$ 2700 \&   1900 $\times$ 1000& 0.03–1 m/pixel & 10000 / 3000 / 3000 & satellite images & building, road \\
SYSU-CD \citep{shi2021deeply} & 20000 & 256 $\times$ 256 & 0.5 m/pixel & 12000 / 4000 / 4000 & aerial images & building, vegetation, road \\
WHU-CD \citep{ji2018fully} & 1 & 32207 $\times$ 15354 & 0.075 m/pixel & 5947 / 743 / 744 & aerial images & building \\
\hline
\end{tabular}}
\vspace{-4mm}
\end{table*}

\subsubsection{Effective Multiscale Feature Fusion Module}
\label{sssec:efmm}
As discussed in Section~\ref{ssec:baseline} that the naive fusion of multiscale features increases the computational complexity of the network while struggling to provide favorable CD performance, therefore, an effective strategy was required for optimal fusion of multiscale representations. We designed the effective multiscale feature fusion (EMFF) module to achieve this objective, as illustrated in Figure~\ref{fig:emff_arch}. Notably, proposed EMFF module does not utilize any learnable parameters layer while being capable of highlighting the critical information within the multiscale representations.
The proposed EMFF module takes the multiscale features $S_j$ and interpolates them to obtain the features having same spatial resolution. Afterwards, it first applies average pooling operation on the features $S_4$ along the channel dimension to obtain the average response ${S'}_4$ within a local neighborhood while making its channel dimension same as the features $S_3$. We empirically found that the average pooling in the deep layers is more effective compared to max pooling operation.
To focus on semantically relevant information, we take the mean of features $S_3$ along channel dimension, apply the \textit{tanh} activation and multiply it with the pooled features ${S'}_4$ to get enhanced features $E_4$. Later we add the representations $E_4$ and $S_3$. Mathematically, these operations are as follows:

\vspace{-1mm}
\begin{equation}
    \begin{array}{l}
    w=tanh(mean_{c}(S_3)) \\
    {S'}_4=AvgPool_c(S_4) \\
    E_4=w\odot{S'}_4 \\
    {E'}_4=S_3+E_4
    \end{array}
\end{equation}
\vspace{-1mm}

After obtaining the enhanced representations ${E'}_4$, we further pass it to the average pooling layer to decrease the channel dimension and obtain the features ${S'}_3$ that depict the mean response within a local neighborhood. 
These features are then added to the representations $S_2$ and fed to the max pooling layer highlighting the significant information and obtaining the features ${S'}_2$. Finally, we add the feature representations $S_1$ and ${S'}_2$ to focus on fine grained information, followed by concatenation with the features $S_4$ to obtain semantically rich representations ($\bar{S}$).

\subsubsection{Training Phase}
During the training phase, the bitemporal images are first passed to the teacher network and obtain the teacher predictions $P_t$. The teacher network is kept frozen during training. 
Subsequently, input images are passed to the student network and obtain the student predictions $P_s$ along with the intermediate output representations $\hat{S}$ as illustrated in Figure~\ref{fig:main_framework}. 

We optimize the entire EFENet framework in an end-to-end fashion using a combination of losses. \textit{Cross-Entropy} loss $\mathcal{L}_{ce}$ is utilized on student predictions $P_s$, encouraging it to align the predictions with ground-truth mask. To further boost the student network's ability to accurately distinguish between the changed and unchanged pixels, we utilize the \textit{Binary Cross-Entropy} loss $\mathcal{L}_{bce}$ on intermediate output $\hat{S}$. Finally, we enable the network to improve its performance by instilling the information contained in the teacher's predictions $P_t$ and utilize the mean absolute error loss $\mathcal{L}_{mae}$ between the teacher and student predictions. The final training objective becomes the combination of three losses as follows:

\vspace{-1mm}
\begin{equation}
    \mathcal{L}_{Total}=\alpha_1\mathcal{L}_{ce} + \alpha_2\mathcal{L}_{bce}+\alpha_3\mathcal{L}_{mae}
\end{equation}
\vspace{-8mm}


\section{Experiments}
\subsection{Implementation Details}
The proposed method is implemented in PyTorch using single A100 GPU, which takes a pair of images of size $256 \times 256$. During training, the pair of images is passed to both teacher and student models. In the training pipeline, we incorporate random flip, color jitter, scale-crop, and Gaussian blur augmentation techniques to enhance the generalization capabilities. We set our teacher network as HyRet-Change \cite{fiaz2025hyret} since it provides promising results on CD task, however, any large network either based on late or early fusion can be utilized as a teacher network.  The student network utilized in experiments is FocalNet-T \cite{yang2022focal} unless otherwise specified in the paper. 
We set batch size as 32, learning rate as $0.0003$ and trained for 600 epochs. We used AdamW optimizer and set weight decay to 0.01, beta values equal to (0.9, 0.999), and used  linear decay to decrease the learning rate to 0 until the last epoch.

\begin{table*}[t!]
\centering
\caption{Performance comparison of the state-of-the-art methods on LEVIR-CD. FLOPs and Latency are computed using RGB image size of $224 \times 224$. 
The proposed early fusion based approach achieves strong balance between the CD performance and efficiency.
Best results are highlighted in bold text and second best are underlined. }
\label{tab:levir_comparison}
\scalebox{0.9}{
\begin{tabular}{l|c|ccc|ccc}
\toprule
\textbf{Method} & \textbf{Backbone} & \textbf{Params} (M) & \textbf{FLOPs} (G) & \textbf{Latency} (msec) & \textbf{IoU} $(\%)$ & \textbf{F1} $(\%)$ & \textbf{Accuracy} $(\%)$ \\
\midrule
EATDer \cite{ma2023eatder} & Custom  & 7.12 & 21.30 & 26.2 & -- & 91.20 & 98.75 \\
ELGCNet-LW \cite{2024_elgcnet} & Custom & 6.78 & 15.17 & 24.5 & 82.36 & 90.33 & 99.03 \\
ChangeFormer \cite{bandara2022changeformer} & Custom & 41.03 & 106 & 26.6 & 82.48 & 90.40 & 99.04 \\
Convformer-CD/48~\cite{yang2025convformer} & Custom & 49.31 & 5.3 & 48.8 & \underline{84.23} & \underline{91.44} & \underline{99.13} \\
USSFC-Net \cite{lei2023ultralightweight} & Custom & \underline{1.52} & 2.53 & 25.9 & -- & 91.04 & -- \\

SFEARNet \cite{li2025sfearnet} & SegFormer & 5.56 & 3.64 & 26.7 & 83.23 & 90.85 & 99.07 \\
TinyCD \cite{codegoni2023tinycd} & EfficientNet & \textbf{0.28} & \textbf{1.12} & 10.4 & 83.57 & 91.05 & 99.10 \\

BIT \cite{bit2022} & ResNet-18 & 12.40 & 8.32 & 13.3 & 80.68 & 89.31 & 98.92 \\
STRobustNet \cite{zhang2025strobustnet} & ResNet-18  & 13.73 & 19.32 & 12.7 & 83.66 & 91.11 & 99.10 \\
TMSF \cite{wang2025novel} & ResNet-18  & 12.92 & 8.90 & 40.4 & 83.29 & 90.88 & -- \\
FSG-Net \cite{xie2025fsg} & ResNet-18  & 13.76 & 5.45 & 18.8 & 83.94 & 91.27 & 99.10 \\

\midrule
\rowcolor{silver}
EFENet (Ours) & mit-b1 & 13.37 & \underline{2.49} & \underline{8.1} & 83.20 & 90.83 & 99.08 \\
\rowcolor{silver}
EFENet (Ours) & ResNet-34 & 21.50 & 4.39 & \textbf{3.8} & 83.33 & 90.91 & 99.09 \\
\rowcolor{silver}
EFENet (Ours) & FocalNet-T & 30.32 & 6.46 & 12.1 & \textbf{84.78} & \textbf{91.76} & \textbf{99.17} \\ 
\bottomrule

\end{tabular}
}
\end{table*}

\subsection{Datasets}
We perform extensive experiments on four publicly available datasets as listed in Table~\ref{tab:cd_datasets}.

\noindent\textbf{LEVIR-CD  \citep{chen2020spatial}}
It is a large-scale bitemporal remote sensing change detection benchmark dataset which consists of 637 pairs of images. Each image size is $1024\times1024$ with 0.5m per pixel spatial resolution. This dataset covers building related changes such as apartments, housing structures, warehouse, and compact garages, over a 5 to 14-year span in multiple cities across Texas, USA. The images are captured from Google Earth and follow the partitioning protocol (7:1:2). The original images are cropped into non-overlapping patches of $256\times256$, yielding 7120 train samples, 1024 validation samples, and 2048 test samples.

\noindent\textbf{CDD-CD  \citep{lebedev2018change}} 
This challenging CD benchmark dataset consists of 11 pair of raw images captured from Google Earth, with seven image pairs of size $4725 \times 2700$ pixels and four image pairs of size $1900 \times 1000$, from 0.03m to 1m per pixel. This is a challenging dataset due to complex scenarios including  seasonal variations, man-made infrastructures (buildings and road infrastructure) and natural features (vegetation/forest cover). This dataset is available in cropped images of size $256\times256$ which results in 16000 image pairs. These cropped images are available in split of 10000, 3000, and 3000 for training, validation, and test purposes, respectively.

\begin{table}[t!]
\centering
\caption{SOTA comparison on WHU-CD benchmark dataset.
Best results are highlighted in bold text and second best are underlined. }
\label{tab:whu_comparison}
\setlength{\tabcolsep}{10pt}
\scalebox{0.8}{
\begin{tabular}{l|ccc}
\toprule
\textbf{Method} & \textbf{IoU} $(\%)$ & \textbf{F1} $(\%)$ & \textbf{Accuracy} $(\%)$ \\
\midrule
BIT \cite{bit2022} & 72.39  & 83.98  & 98.75  \\
ChangeFormer \cite{bandara2022changeformer} & 73.80  & 84.93 & 98.82  \\
ELGCNet \cite{2024_elgcnet} & 80.86 & 89.42 & 99.20 \\
ScratchFormer \cite{noman2024remote} & 84.97 & 91.87 & 99.37 \\
EATDer \cite{ma2023eatder}  & -- & 90.01 & 98.58 \\
STRobustNet \cite{zhang2025strobustnet} & 83.29 & 90.89 & 99.32  \\
ConvFormer-CD \cite{yang2025convformer} & 85.41  & 92.13 & 99.26  \\
TinyCD \cite{codegoni2023tinycd} & 84.74 & 91.74 & 99.34 \\
TMSF \cite{wang2025novel} & 80.09 & 88.95 & -- \\
SFEARNet \cite{li2025sfearnet}  & \underline{85.81} & \underline{92.36} & \underline{99.38} \\
\midrule
\rowcolor{silver}
EFENet (Ours) & \textbf{87.17} & \textbf{93.15} & \textbf{99.47} \\
\bottomrule
\end{tabular}}
\vspace{-2mm}
\end{table}

\noindent\textbf{SYSU-CD  \citep{shi2021deeply}} 
This dataset comprises of 20000 ultrahigh resolution images of size $256\times256$ and the ground sample distance resolution of 0.5m,  acquired between 2007 and 2014 in Hong Kong region. This dataset pose various land-cover change types, including vegetation changes, road/transportation expansions, newly constructed urban buildings, sea construction, and pre-construction ground works. For experimental comparison, we utilize standard split provided, which has 12000 pairs of images for training and 4000 pairs for validation and testing, corresponding to a ratio of 6:2:2.

\noindent\textbf{WHU-CD  \citep{ji2018fully}} 
WHU building dataset is another aerial building-related CD benchmark dataset which consists of a high-resolution (0.075m) image pair of size $32207 \times 15354$ pixels. This dataset is challenging due to the variety of building infrastructures, such as countryside, residential, and industrial areas, with different sizes and colors. The data set is cropped into non-overlapping regions of size $256\times256$, and is divided into 5947, 743, and 744 image pairs for training, validation, and test sets, respectively. 

\subsection{Evaluation Protocol} Following prior works \cite{bandara2022changeformer, bit2022, 2024_elgcnet, zhang2025strobustnet}, we utilize the change-class intersection over union (IoU), change-class F1 score, and overall accuracy to measure the performance of the model. In addition, we utilize learnable parameters, floating-point operations per inference (FLOPs) and latency to evaluate the computational complexity of the model. For fair comparison, we utilize the publicly available official repositories of the compared methods to verify the number of parameters and FLOPs. Additionally, we use the Quadro RTX 6000 GPU to compute the latency of all methods.

\begin{table}[t!]
\centering
\caption{SOTA comparison on CDD-CD dataset. 
Best results are highlighted in bold text and second best are underlined. }
\label{tab:cdd_comparison}
\setlength{\tabcolsep}{10pt}
\scalebox{0.8}{
\begin{tabular}{l|ccc}
\toprule
\textbf{Method} & \textbf{IoU} $(\%)$ & \textbf{F1} $(\%)$ & \textbf{Accuracy} $(\%)$ \\
\midrule
BIT \cite{bit2022} & 80.01 & 88.90 & 97.47 \\
ChangeFormer \cite{bandara2022changeformer} & 81.53 & 89.83 & 97.68 \\
STRobustNet \cite{zhang2025strobustnet} & 88.08 & 93.66 & 98.50 \\
ConvFormer-CD \cite{yang2025convformer} & 88.63 & 93.96 & 98.59 \\
USSFC-Net \cite{lei2023ultralightweight} & -- & 94.74 & -- \\
DSFDcd \cite{wang2025dsfdcd} & 88.81 & 94.06 & 98.57 \\
FSG-Net \cite{xie2025fsg} & 88.96 & 94.16 & 98.56 \\
TMSF \cite{wang2025novel} & 90.44 & 94.98 & -- \\
EATDer \cite{ma2023eatder} & -- & 95.97 & 98.97 \\
ELGCNet-LW \cite{2024_elgcnet} & \underline{93.48} & \underline{96.63} & \underline{99.21} \\
\midrule
\rowcolor{silver}
EFENet (Ours) &  \textbf{94.83} & \textbf{97.34} & \textbf{99.37} \\
\bottomrule
\end{tabular}}
\vspace{-2mm}
\end{table}

\subsection{Quantitative Comparison}
Here, we discuss the quantitative comparison of our approach with state-of-the-art (SOTA) methods over LEVIR-CD, CDD-CD, SYSU-CD, and WHU-CD benchmark datasets.

\noindent\textbf{LEVIR-CD:} We compare the performance of the proposed EFENet with SOTA methods on the LEVIR-CD dataset in Table~\ref{tab:levir_comparison}. We observe that recent Convformer-CD/48 \cite{yang2025convformer} provides promising results by achieving the IoU score of 84.23\% and FLOPs of 5.3G while having high latency and large number of  parameters i.e., 48.8 msec and 49.31M, respectively. 
Similarly, FSG-Net \cite{xie2025fsg} is another SOTA method having latency of 18.8 msec while achieving IoU score of 83.94\%.  
Although TinyCD \cite{codegoni2023tinycd} that utilizes a light-weight backbone, provides fewer parameters and FLOPs, it has reduced IoU score of 83.57\% despite utilizing a sophisticated decoder and feature fusion modules. 
Other notable lightweight methods include STRobustNet \cite{zhang2025strobustnet}, SFEARNet \cite{li2025sfearnet}, and TMSF \cite{wang2025novel} obtaining the IoU score of 83.66\%, 83.23\%, and 83.29\%, respectively, while possessing a high latency. In contrast, our method provides a favorable performance-efficiency tradeoff by obtaining an IoU score of 84.78\%  while exhibiting latency of 12.1 msec and 6.46 GFLOPs, which demonstrates the merits of our approach.

\begin{figure*}[t!]
\centering
\includegraphics[width=1.0\linewidth]{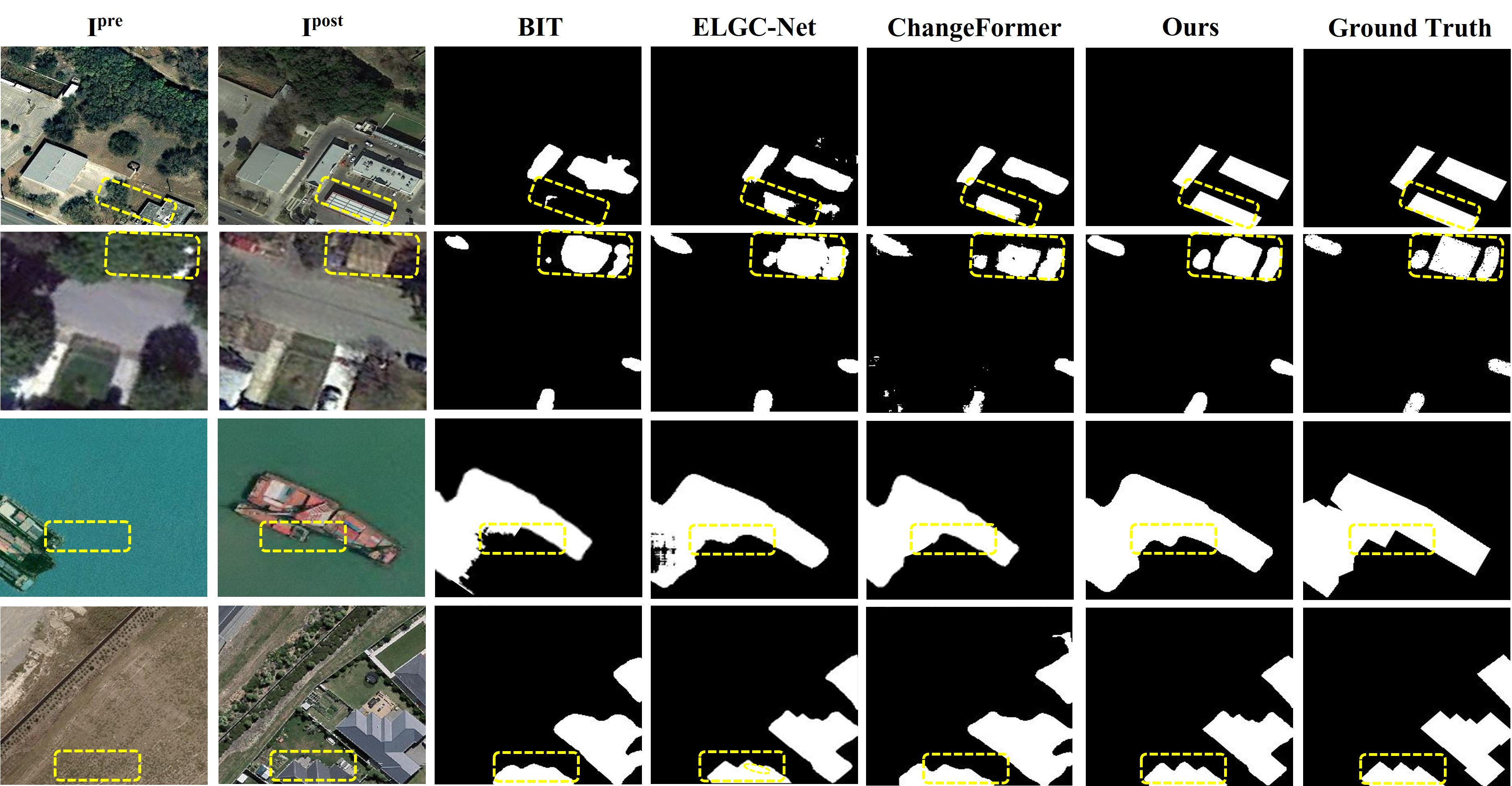} 
\caption{Qualitative comparison of EFENet with BIT \cite{bit2022}, ChangeFormer \cite{bandara2022changeformer}, and ELGC-Net \cite{2024_elgcnet} methods. Data samples shown from row one to four correspond to LEVIR-CD, CDD-CD, SYSU-CD, and WHU-CD datasets, respectively. Notably, our approach demonstrates its capabilities to better detect the semantic changes highlighted in the yellow dotted boxes compared to existing methods.}
\label{fig:qualitative_comparison}
\end{figure*}

\begin{table}[t]
\centering
\caption{SOTA comparison on SYSU-CD benchmark dataset. 
Best results are highlighted in bold text and second best are underlined. }
\label{tab:sysu_comparison}
\setlength{\tabcolsep}{10pt}
\scalebox{0.8}{
\begin{tabular}{l|ccc}
\toprule
\textbf{Method} & \textbf{IoU} $(\%)$ & \textbf{F1} $(\%)$ & \textbf{Accuracy} $(\%)$ \\
\midrule
ChangeFormer \cite{bandara2022changeformer} & 60.60 & 75.46 & 89.20 \\
BIT \cite{bit2022} & 61.40 & 76.08 & 88.95 \\
ELGCNet \cite{2024_elgcnet} & 66.62 & 79.97 & 90.72 \\
STRobustNet \cite{zhang2025strobustnet} & 67.59 & 80.66 & 91.13 \\
ConvFormer-CD/48 \cite{yang2025convformer} & 65.76 & 79.35 & 90.98 \\
EATDer \cite{ma2023eatder}  & 66.99 & 80.23 & 90.34 \\
DSFDcd \cite{wang2025dsfdcd} & 67.31 & 80.46 & 91.00 \\
RHighNet \cite{dong2025relation} & \underline{67.53} & \underline{80.62} & \underline{91.33} \\
\midrule
\rowcolor{silver}
EFENet (Ours) & \textbf{68.67}  & \textbf{81.42} & \textbf{91.67} \\
\bottomrule
\end{tabular}
}
\end{table}

\noindent\textbf{WHU-CD:} Table \ref{tab:whu_comparison} further illustrates the performance comparison of our approach with widely adopted SOTA methods over WHU-CD dataset. We notice that BIT \cite{bit2022} and ChangeFormer \cite{bandara2022changeformer} struggle a lot on this dataset. Recent prominent methods such as STRobustNet \cite{zhang2025strobustnet} and TinyCD \cite{codegoni2023tinycd} obtain IoU scores of 83.29\% and 84.74\%, whereas our approach demonstrates a significant gain of 3.88\% and 2.43\%, respectively. In addition, ConvFormer-CD \cite{yang2025convformer} and SFEARNet \cite{li2025sfearnet} perform better and present IoU score of 85.41\% and 85.81\%. Our method obtains 87.17\% IoU surpassing these methods and attains an absolute gain of 1.76\% and 1.36\%, respectively. These results demonstrate the elegance of EFENet by achieving a consistent improvement across all metrics.

\noindent\textbf{CDD-CD:} In Table \ref{tab:cdd_comparison}, we compare our approach with existing SOTA methods over CDD-CD. It is noticeable that TMSF \cite{wang2025novel}, EATDer \cite{ma2023eatder}, and ELGCNet-LW \cite{2024_elgcnet} attain F1 score of 94.98\%, 95.97\%, and 96.63\%. Whereas all other approaches exhibit F1 score less than 95\%. On this challenging dataset, our approach surpasses other methods and shows consistent performance over IoU, F1, and OA metrics obtaining scores of 94.83\%, 97.34\%, and 99.37\%, respectively.

\noindent\textbf{SYSU-CD:} We also evaluate our approach on the SYSU-CD benchmark dataset in Table \ref{tab:sysu_comparison}. We observe that BIT \cite{bit2022} and ChangeFormer \cite{bandara2022changeformer} struggle on this data set and achieve an IoU of 61.40\% and 60.60\%.  Our approach attains a gain of 7.27\% and 8.07\% over these well known CD methods. However, compared to recent models including STRobustNet \cite{zhang2025strobustnet}, RHighNet \cite{dong2025relation}, and DSFDcd \cite{wang2025dsfdcd}, our method shows consistent improvement in terms of IoU, F1, and overall accuracy metrics and demonstrates favorable performance.

\subsection{Qualitative Comparison}
In Figure \ref{fig:qualitative_comparison}, we present a qualitative comparison of our approach with the well known change detection methods including BIT \cite{bit2022}, ELGC-Net \cite{2024_elgcnet}, and ChangeFormer \cite{bandara2022changeformer}. We notice that EFENet is capable of capturing semantic changes at multiple scales in complex scenarios. For instance, in first and second rows corresponding to LEVIR-CD and CDD-CD respectively, our method detects the semantic change regions fairly better compared to other methods. Furthermore, illustrated images in third and fourth rows corresponding to SYSU-CD and WHU-CD, our method performs favorably whereas the existing methods struggle to accurately detect the boundaries of the change regions on these complex scenes. 
These results demonstrate the strong capabilities of EFENet by precisely detecting the relevant semantic changes despite using the early fusion approach.

\subsection{Ablation Experiments}
\noindent\textbf{EFENet Components: } Table~\ref{tab:ablation_components} presents the effectiveness of the proposed components of EFENet. We observe that the baseline network provides unsatisfactory performance by achieving IoU score of 80.83\% while suffering from greater number of FLOPs. When EMFF replaces the convolution-based multiscale feature fusion in baseline network, CD performance is improved by achieving better IoU score of 82.08\% while parameters and GFLOPs are reduced to 30.32 and 6.46, respectively. The performance of the network is further improved when we utilize $\mathcal{L}_{bce}$ on the intermediate output features $\bar{S}$. Finally, the proposed EFENet (row 4 in Table~\ref{tab:ablation_components}) framework provides significant performance improvement obtaining IoU score of 84.78\%. 

\begin{table}[t!]
\centering
\caption{Ablation study demonstrating the effectiveness of the proposed components on LEVIR-CD utilizing FocalNet-T \cite{yang2022focal} backbone. }
\label{tab:ablation_components}
\scalebox{0.58}{
\setlength{\tabcolsep}{4pt}
\begin{tabular}{l|cc|ccc}
\toprule

\textbf{Method} & \textbf{Params} (M) & \textbf{FLOPs} (G) &  \textbf{IoU} $(\%)$ & \textbf{F1} $(\%)$ & \textbf{Accuracy} $(\%)$ \\
\midrule
Baseline & 31.23 & 9.30 & 80.83 & 89.40  & 98.97 \\
Baseline + EMFF & 30.32  & 6.46 & 82.08  & 90.15  & 99.05  \\
Baseline + EMFF + $\mathcal{L}_{bce}$ & 30.32  & 6.46 & 82.77  & 90.57  & 99.19  \\
Baseline + EMFF + $\mathcal{L}_{bce}$ + $\mathcal{L}_{mae}$ & 30.32  & 6.46 & 84.78 & 91.76 & 99.17 \\
\bottomrule
\end{tabular}
}
\end{table}

\noindent\textbf{Ablation of Losses: } In Table~\ref{tab:ablation_losses}, we demonstrate the impact of different losses on the performance of the proposed EFENet. From the table, we observe that the Cross Entropy (CE) loss ($\mathcal{L}_{ce}$ in Figure~\ref{fig:main_framework}) when combined with mean absolute error (MAE) loss provides the best IoU score of 84.78\%. The replacement of CE loss with mIoU loss slightly degrades the performance to 84.56\%. Whereas using KL divergence loss or mean squared error loss in place of MAE loss results in noticeable performance reduction as illustrated in first two rows of the table.

\begin{table}[t!]
\centering
\caption{Demonstration of various losses utilized for information distillation on LEVIR-CD utilizing FocalNet-T \cite{yang2022focal} backbone. }
\label{tab:ablation_losses}
\scalebox{0.65}{
\setlength{\tabcolsep}{9pt}
\begin{tabular}{l|cc|ccc}
\toprule
\textbf{Method} & $\bm{Loss(P_s)}$ & $\bm{Loss(P_s,P_t)}$ & \textbf{IoU} $(\%)$ & \textbf{F1} $(\%)$ & \textbf{Accuracy} $(\%)$ \\
\midrule
\multirow{4}{*}{EFENet} & CE & KL Div & 82.53  & 90.43  & 99.05 \\
 & CE & MSE & 83.84  & 91.21  & 99.11 \\
 & mIoU & MAE & 84.56  & 91.63 & 99.15 \\
 & CE & MAE & 84.78 & 91.76 & 99.17 \\
\bottomrule
\end{tabular}
}
\end{table}

\noindent\textbf{How various encoders affect the performance: } We further investigate the effect of different encoder architectures on the performance of the EFENet. From Table~\ref{tab:ablation_backbone}, we notice that the CNN based encoder such as ResNet-34 provides optimal tradeoff between CD performance and speed by achieving IoU score of 83.33\% and latency of 3.8 msec. 
Whereas the utilization of transformer-based backbone such as mit-b0 \cite{xie2021segformer} provides acceptable results by achieving IoU score of 81.23\% while its latency is increased to 7.9 msec despite of having less number of FLOPs as compared to ResNet-34. 
We observe similar trend between the FocalNet-T and DebiFormer-T encoders. From these results, we infer that the CNN based encoder maintain better balance between the CD performance and latency despite of having greater number of parameters and FLOPs compared to transformer based encoders.

\begin{table}[t!]
\centering
\caption{Performance comparison of various backbone networks utilized in the EFENet framework on LEVIR-CD. }
\label{tab:ablation_backbone}
\scalebox{0.5}{
\setlength{\tabcolsep}{8pt}
\begin{tabular}{l|ccc|ccc}
\toprule
\textbf{Backbone} & \textbf{Params} (M) & \textbf{FLOPs} (G) & \textbf{Latency} (msec) & \textbf{IoU} $(\%)$ & \textbf{F1} $(\%)$ & \textbf{Accuracy} $(\%)$ \\
\midrule
mit-b0 \cite{xie2021segformer} & 3.37  & 0.69 & 7.9 & 81.23  & 89.45  & 98.96  \\
mit-b1 \cite{xie2021segformer} & 13.37  & 2.49 & 8.1 & 83.20  & 90.83  & 99.08  \\
ResNet-34 \cite{he2016resnet} & 21.50 & 4.39 & 3.8 & 83.33  & 90.91  & 99.09  \\
Debiformer-T \cite{baolong2024debiformer} & 21.19  & 3.79 & 31.2  & 83.63  & 91.09  & 99.11  \\
FocalNet-T \cite{yang2022focal} & 30.32 & 6.46 & 12.1  & 84.78 & 91.76 & 99.17 \\
\bottomrule
\end{tabular}
}
\end{table}

\section{Conclusion}
This work introduces a simple yet effective framework, named EFENet, for change detection task. The proposed framework is based on early fusion of temporal features and replaces the decoder with a parameter-free feature fusion module (EMFF) demonstrating that early fusion networks are capable of achieving favorable performance while considerably reducing the computational overhead. 
Proposed EFENet further reveals that the performance of change detection varies significantly by using different encoder networks.
Furthermore, we conclude that parameter-free feature fusion is capable of emphasizing the significant information necessary for change detection while maintaining a balance between performance and efficiency.

{
    \small
    \bibliographystyle{ieeenat_fullname}
    \bibliography{main}
}

\end{document}